\documentclass[twocolumn, switch]{article} 
\usepackage{authblk}

\usepackage{preprint}

\usepackage{amsmath, amsthm, amssymb, amsfonts}

\usepackage[numbers,square]{natbib}
\usepackage{hyperref}	

\usepackage{titlesec}
\titlespacing\section{0pt}{12pt plus 3pt minus 3pt}{1pt plus 1pt minus 1pt}
\titlespacing\subsection{0pt}{10pt plus 3pt minus 3pt}{1pt plus 1pt minus 1pt}
\titlespacing\subsubsection{0pt}{8pt plus 3pt minus 3pt}{1pt plus 1pt minus 1pt}

\usepackage{times}
\usepackage{latexsym}
\usepackage{graphicx}
\usepackage{balance}
\usepackage{amsmath}
\usepackage{comment}
\usepackage[boxruled,vlined]{algorithm2e}
\usepackage{comment}
\usepackage{booktabs}
\usepackage{colortbl}
\usepackage[margin=0cm]{caption}
\usepackage[font=small,labelfont=bf]{caption}
\usepackage{amssymb}
 
\usepackage{microtype}
\newcommand\SLASH{\char`\\} 

\usepackage{footnote}
 
 \DeclareMathOperator*{\argmin}{arg\,min}
 
\usepackage{caption}

\SetCommentSty{mycommfont}

\SetKwInput{KwInput}{Input}                
\SetKwInput{KwOutput}{Output}              
\def\BibTeX{{\rm B\kern-.05em{\sci\kern-.025em b}\kern-.08emT\kern-.1667em\lower.7ex\hbox{E}\kern-.125emX}}
 
\usepackage{silence}
\vbadness=\maxdimen
\newlength\mylen
\newcommand\myinput[1]{%
  \settowidth\mylen{\KwIn{}}%
  \setlength\hangindent{\mylen}%
  \hspace*{\mylen}#1\\}
\SetEndCharOfAlgoLine{}

\usepackage{microtype}

\title{Generating Fact Checking Summaries for Web Claims}

\author[1]{\textbf{Rahul Mishra}}
\author[2]{\textbf{Dhruv Gupta}}
\author[3]{\textbf{Markus Leippold}}
\affil[1]{University of Stavanger, Norway}
\affil[1]{\texttt{rahul.mishra@uis.no}}
\affil[2]{Max Planck Institute for Informatics, Germany}
\affil[2]{\texttt{dhgupta@mpi-inf.mpg.de}}
\affil[3]{University of Zurich, Switzerland}
\affil[3]{\texttt{markus.leippold@bf.uzh.ch}}
\begin{document}

	\twocolumn[ 
\begin{@twocolumnfalse} 
	
	\maketitle
	
	\begin{abstract}
We present SUMO, a neural attention-based approach that learns to establish the
correctness of textual claims based on evidence in the form of text documents
(e.g., news articles or Web documents). SUMO further generates an extractive
summary by presenting a diversified set of sentences from the documents that
explain its decision on the correctness of the textual claim. Prior approaches
to address the problem of fact checking and evidence extraction have relied on
simple concatenation of claim and document word embeddings as an input to
claim driven attention weight computation. This is done so as to extract
salient words and sentences from the documents that help establish the
correctness of the claim. However, this design of claim-driven attention does not
capture the contextual information in documents properly. We improve on the
prior art by using improved claim and title guided hierarchical attention
to model effective contextual cues. We show the efficacy of our approach on datasets concerning political, healthcare, and environmental issues.

	\end{abstract}
	\vspace{0.35cm}
	
\end{@twocolumnfalse} 
] 

\section{Introduction}\label{sec:1}
\begin{figure*}[!htbp]
          \centering
		  \includegraphics[width=2\columnwidth]{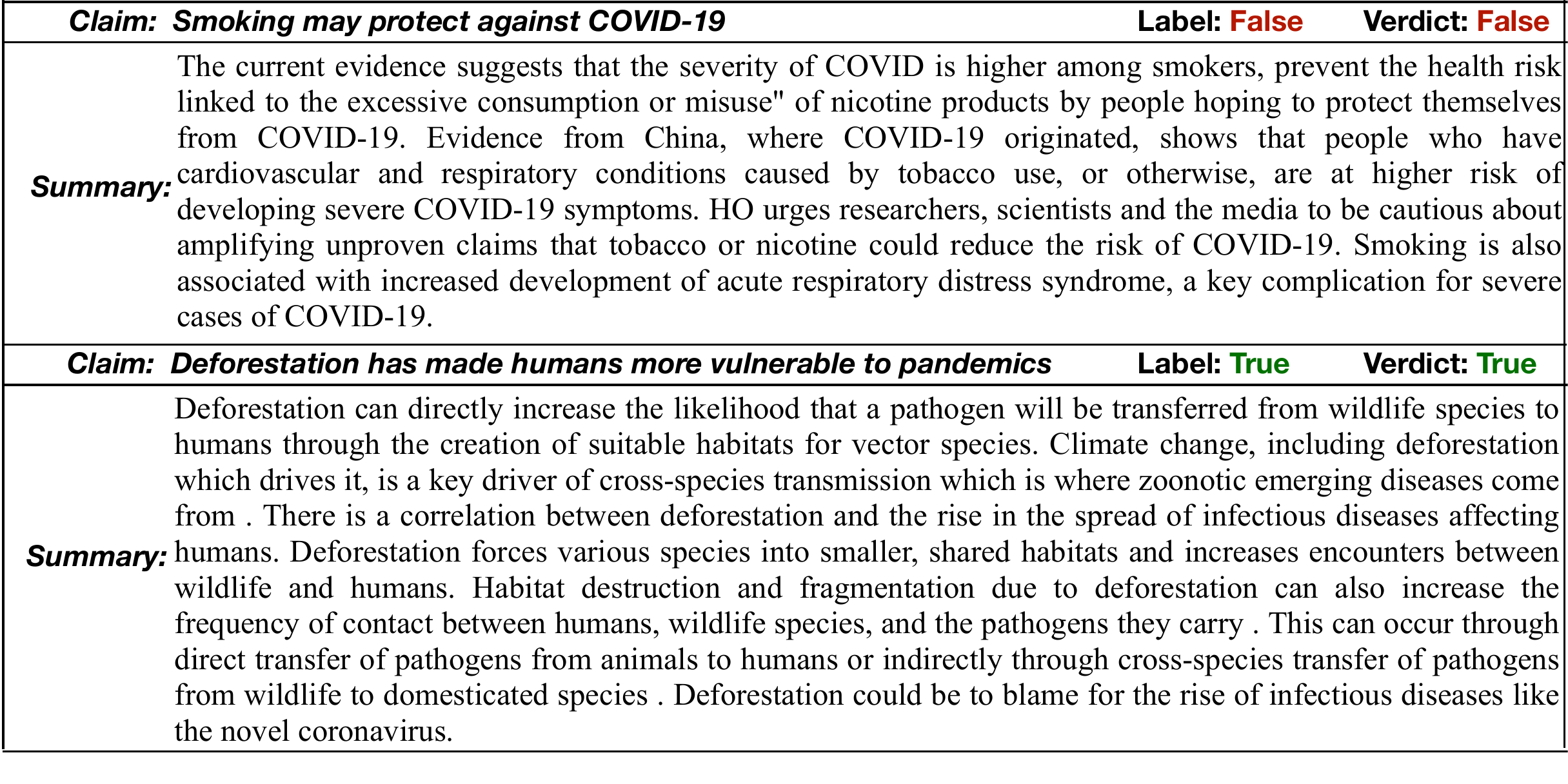}
          \caption{Example summaries generated by \textsc{sumo} for unverified claims on the Web.}
          \label{fig:summary}
\end{figure*}

Most of the information consumed by the world is in the form of digital news, blogs, and social media posts available on the Web. However, most of this information is written in the absence of facts and evidences.
Our ever-increasing reliance on information from the Web is becoming a severe problem as we base our personal decisions relating to politics, environment, and health on unverified information available online. For example, consider the following unverified claim on the Web:
\begin{quote}
	\begin{center}
	\emph{\texttt{"Smoking may protect against COVID-19."}} 
	\end{center}
\end{quote}
A user attempting to verify the correctness of the above claim will often take the following steps: issue keyword queries to search engines for the claim; going through the top reliable news articles; and finally making an informed decision based on the gathered information. Clearly, this approach is laborious, takes time, and is error-prone. In this work, we present \textsc{sumo}, a neural approach that assists the user in establishing the correctness of claims by automatically generating explainable summaries for fact checking. Example summaries generated by \textsc{sumo} for couple of Web claims are given in Figure~\ref{fig:summary}.

\textbf{Prior approaches} to automatic fact checking rely on predicting the credibility of facts \cite{popat2017truth}, instance detection \cite{Ma18,Xu2018}, and fact entailment in supporting documents~\cite{Parikh2016}. The majority of these methods rely on linguistic features \cite{popat2017truth,potthast2018stylometric,Qazvinian}, social contexts, or user responses \cite{Ma15} and comments. However, these approaches do not help explain the decisions generated by the machine learning models. Recent works such as \cite{Atanasova2019,sadhan9,popat2018declare} overcome the explainability gap by extracting snippets from text documents that support or refute the claim. \cite{sadhan9,popat2018declare}~apply claim-based and latent aspect-based attention to model the context of text documents. \cite{sadhan9} model latent aspects such as the speaker or author of the claim, topic of the claim, and domains of retrieved Web documents for the claim. We observe in our experiments that in prior works \cite{sadhan9,popat2018declare}, the design of claim guided attention in these methods is not effective and latent aspects such as the topic and speaker of claims are not always available. The snippets extracted by such models are not comprehensive or topically diverse. To overcome these limitations, we propose a novel design of claim and document title driven attention, which better captures the contextual cues in relation to the claim. In addition to this, we propose an approach for generating summaries for fact-checking that are non-redundant and topically diverse. 

\textbf{Contributions.} Contributions made in this work are as follows. First, we introduce \textsc{sumo}, a method that improves upon the previously used claim guided attention to model effective contextual representation. Second, we propose a novel attention on top of attention (Atop) method to improve the overall attention effectiveness. Third, we present an approach to generate topically diverse multi-document summaries, which help in explaining the decision \textsc{sumo} makes for establishing the correctness of claims. Fourth, we provide a novel testbed for the task of fact checking in the domain of climate change and health care.

\textbf{Outline.} The outline for the rest of the article is as follows. In Section~\ref{sec:related-work}, we describe prior work in relation to our problem setting. In Section~\ref{sec:approach}, we formalize the problem definition 
and describe our approach, \textsc{sumo}, to generate explainable summaries for fact checking of textual claims. In Sections~\ref{sec:datasets} and~\ref{sec:results}, we describe the experimental setup that includes a description of the novel datasets that we make available to the research community and an analysis of the results we have obtained. In Section~\ref{sec:conclude}, we present the concluding remarks of our study.

\section{Related work}\label{sec:related-work}


We now describe prior work related to our problem setting. First, we describe works that rely only on features derived from documents that support the input textual claim. Second, we describe works that additionally include features derived from social media posts in connection to the claim. Third and finally, we describe works that rely on extracting textual snippets from text documents to explain a model's decision on the claim's correctness. 

\subsection{Content Based Approaches} 
Prior approaches for fact checking vary from simple machine learning methods
such as SVM and decision trees to highly sophisticated deep learning methods.
These works largely utilize features that model the  linguistic and stylistic
content of the facts to learn a classifier
\cite{Castillo2011InformationCO,Ma_ijcai,Qazvinian,rashkin-etal-2017-truth}.
The key shortcomings of these approaches are as follows. First, classifiers
trained on linguistic and stylistic features perform poorly as they can
be misguided by the writing style of the false claims, which are deliberately made to look similar to true claims but are factually
false. Second, these methods lack in terms of user response and social context pertaining to the claims, which is very helpful in establishing the
correctness of facts. 

\subsection{Social Media Based Approaches} 
Works such as~\cite{Feng_ijcai2018,Shuwsdm,YangAAAI} overcome the issue of user feedback by using
a combination of content-based and context-based  features derived
from related social media posts. Specifically, the features derived  from
social media include propagation patterns of claim related posts on social
media and user responses in the form of replies, likes, sentiments, and
shares. These methods outperform content-based methods significantly. In
\cite{YangAAAI}, the authors propose a probabilistic graphical model for
causal mappings among the post's credibility, user's opinions,  and user's
credibility.   In \cite{Feng_ijcai2018}, the authors introduce a user response generator based on a deep neural network that leverages the user's past actions such as
comments, replies, and posts to generate a synthetic response for new social
media posts. 

\subsection{Model Explainability} 
Explaining a machine learning model's decision
is becoming an important problem. This is because modern neural network based
methods are increasingly being used as black-boxes. There exist few machine learning models for fact checking that explain this decision via summaries. Related works~\cite{sadhan9,popat2018declare} achieve significant improvement in
establishing the credibility of textual claims by using external evidences from
the Web. They additionally extract snippets from evidences that explain their
model's decision. However, we find that the  claim-driven attention design used in
these methods is inadequate, and does not capture sufficient  context of the
documents in relation to the input claim. The snippets extracted by these
methods are often redundant and lack topical diversity offered by Web
evidences. In contrast, our method enhances the claim-driven attention
mechanism and generates a topically diverse, coherent multi-document
summary for explaining the correctness of claims.
\section{SUMO}\label{sec:approach}

We now formally describe the task of fact checking and explain \textsc{sumo} in detail. \textsc{sumo} works in two stages. In the first stage, it predicts the correctness of the claim. In the second stage,  it generates a topically diverse summary for the claims. As input, we are provided with a Web claim $ c\in C$, where $C$ is a collection of Web claims and a pseudo-relevant set of documents 
$D = \{d_1, d_2, \ldots , d_{m}\}$, where $m$ is the number of results retrieved for claim $c$. The documents $d \in D$ are retrieved from the Web as potential evidences, using claim $ c$ as a query. Each retrieved document $d$ is accompanied by its title $t$ and text body $bd$, i.e. ($d = \langle t, bd\rangle$). We define the representation of each document's body as a collection of $k$ sentences as $bd = \{s_1, s_2, ..., s_k\} $ and each sentence as the collection of $l$ words as $\{w_1, w_2, ..., w_l\} \in \mathbb{W}$, where $\mathbb{W}$ is the overall word vocabulary of the corpus. By $k$ and $l$, we denote the maximum numbers of sentences in a document and the maximum number of words in a sentence, respectively.  We use both \textsc{word2vec} and pre-trained GloVe embeddings to obtain the vector representations for each claim, title, and document body.  The objective is to classify the claim as either true or false and automatically generate a topically diverse summary pieced together from $D$ for establishing the correctness of the claim.

\begin{figure*}[t]
          \centering
          \includegraphics[width=2.1\columnwidth]{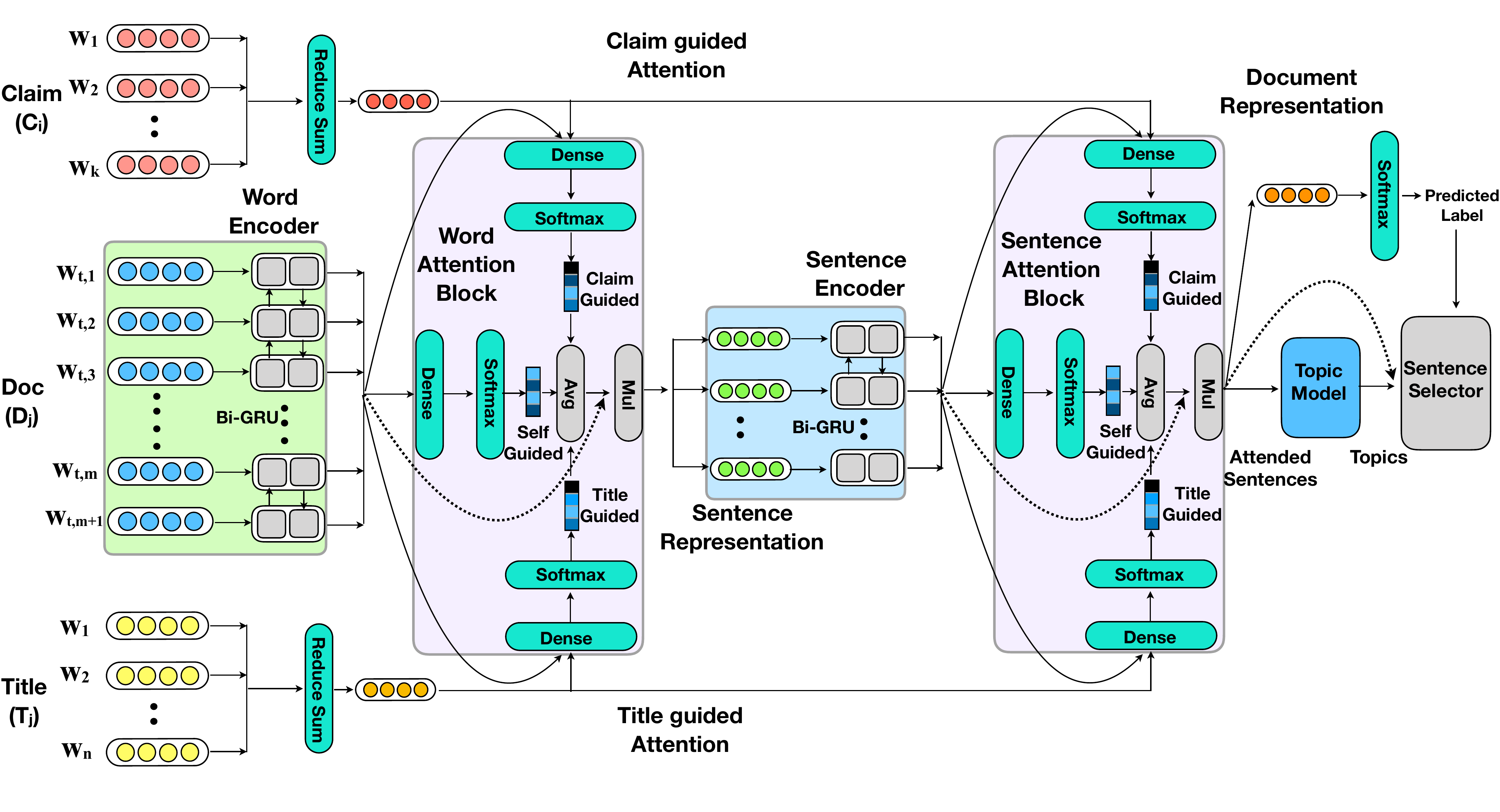}
          \caption{\textsc{sumo}'s neural network architecture for establishing the correctness of Web claims.}
          \label{fig:attention}
\end{figure*}

\subsection{Predicting Claim Correctness by Neural Attention}

We now describe \textsc{sumo}'s neural architecture (see
Figure~\ref{fig:attention}) that helps in predicting the correctness of the
input claim along with its pseudo-relevant set of documents. The model
additionally learns the weights to words and sentences in the document's body
that help ascertain the claim's correctness.  First, we need to encode
the pseudo-relevant documents that support a claim. To this end, as a \textbf{sequence
encoder}, we use a Gated Recurrent Unit (GRU) to encode the document's body
content. 
Claim and document's title are not encoded using sequence encoder; we explain the method to represent them in detail in upcoming sections.

\textbf{Claim-driven Hierarchical Attention.}, aims to attend salient words that are significant and
have relevance to the content of the claim. Similarly, we aim to
attend the salient sentences at the sentence level attention. Recent
works have used claim guided attention to model the contextual representation
of the retrieved documents from the Web. 
These approaches provide claim-guided attention by first concatenating the claim word embeddings with document word embeddings and then applying a dense softmax layer to learn the
attention weights as follows:
\begin{equation}\begin{aligned}{r}_i = c_i\mathbin\Vert d_i\text{~~\&~~}a_i = \tanh(W_a{r}_i+ b_a)\\
  \alpha = \textrm{softmax}(a_i),\label{old_att}\end{aligned}\end{equation}
where $c_i$ and $d_i$ are the $i^{th}$ claim and document embeddings. $W_a$ and $b_a$ are the weight matrix and bias and
$\alpha$ is the learned attention weight. However, during
experiments, we observe that applying claim-based
attention provides an inferior overall
document representation. Therefore, we do not concatenate the claim and document
embeddings before attention weight computation. 

Each claim $c_i$ is consists
of $l$ maximum number of words as $\{w_1,w_2,......,w_l\}$. We represent each
claim $c_i$ as the summation of embeddings of all the words contained in it as: $Cl_i = \sum_{j=1}^{l}f(w_j)$, where $f(w_j)$ is the word embedding of the $j_{th}$ word of claim $c_i$.
Claim representation $Cl_i$ and hidden states
$h_j$ from the GRU are used to \textbf{compute word-level claim-driven attention weights} as:
\begin{equation}
  \begin{aligned}
  u_{j,i} = \tanh(W_{j,i} h_j + b_{j,i}) \\
  \alpha^C_{j,i} = \textrm{softmax}(u^\top_{j,i} Cl_i),
  \end{aligned}
\end{equation}
where $W_{j,i}$ and $ b_{j,i}$ are the weight matrix and bias,
$\alpha^C_{j,i}$ is the word level claim driven attention weight vector, and
$h_j = (h_{j,1}, h_{j,2}, ..., h_{j,l})^\top$ represents the tuple of all the
hidden states of the words contained in the $j^{th}$ sentence. To
\textbf{compute sentence level claim-driven attention weights}, we use claim
representation $Cl_i$ and hidden states $h^S_j$ from the sentence level GRU
units as concatenations of both forward and backward hidden states $h^S_{j} =
\overrightarrow{h^S_{j}}\mathbin\Vert \overleftarrow{h^S_{j}}$ as follows:
\begin{equation}\begin{aligned}
u_{j} = \tanh(W_{j} h^S + b_{j,i}) \\
\alpha^C_{j} = \textrm{softmax}(u^\top_{j} Cl_i),
\end{aligned}
\end{equation}
where $W_{j}$ and $ b_{j}$ are the weight matrix and bias, $h^S = (h^S_{1}, h^S_{2}, ..., h^S_{l})^\top$ is the combination of all hidden states from sentences, and $\alpha^C_{j} = (\alpha_{j,1}, \alpha_{j,2}, ...,\alpha_{j,k})^\top$ is the sentence level claim-driven attention weight vector for the $j^{th}$ document.

\textbf{Title-driven Hierarchical Attention.} The objective of using the
document title is to guide the attention in capturing sections in the document that are more critical and
relevant for the title. Articles convey multiple perspectives, often reflected in  their titles. By title-driven attention, we attend to those words and sentences that are not covered in claim-driven
attention. Title-driven attention at both word and sentence level can be computed in a similar fashion as claim-driven attention.  Each title $t_i$ is comprised of $l$ maximum number of
words as $\{w_1,w_2, \ldots,w_l\}$. We represent each claim $t_i$ as the
summation of embeddings of all the words contained in it as: $T_i =
\sum_{j=1}^{l}f(w_j)$. Title-driven attention weights for both words and
sentence level can be computed as follows:\begin{equation}\begin{aligned}u_{j,i} = \tanh(W_{j,i} h_j + b_{j,i}) \\\alpha^T_{j,i} = \textrm{softmax}(u^\top_{j,i} T_i)\\u_{j} = \tanh(W_{j} h^S + b_{j,i}) \\\alpha^T_{j} =\textrm{softmax}(u^\top_{j} T_i).\end{aligned}\end{equation}
\textbf{Hierarchical Self-Attention.} Self-attention is a simplistic form of
attention. It tries to attend salient words in a sequence of words and
salient sentences in a collection of sentences based on the self context of a
sequence of words or a collection of sentences. In addition to claim-driven and
title-driven attention, we apply self-attention to capture the unattended words
and sentences which are not related to claim or title directly but are very
useful for classification and summarization. Self-attention weights for both
words and sentence level can be computed as follows:\vspace{-2mm} \begin{equation}\begin{aligned}u_{j,i} = \tanh(W_{j,i} h_j + b_{j,i}) \\\alpha^{Sl}_{j,i} = \textrm{softmax}(u^\top_{j,i})\\
u_{j} = \tanh(W_{j} h^S + b_{j,i})\\\alpha^{Sl}_{j} = \textrm{softmax}(u^\top_{j}),\end{aligned}\end{equation} where $\alpha^{Sl}_{j,i}$ and $\alpha^{Sl}_{j} $ are the self-attention weight vectors at word and sentence levels respectively.

\textbf{Fusion of Attention Weights.} We combine the attention weights from the three kinds of attention mechanisms: claim-driven, title-driven, and self-attention at both the word and sentence levels. At the word level, we set:
\begin{equation}\begin{aligned}\alpha_j = (\alpha^C_{j,i} + \alpha^T_{j,i} + \alpha^{Sl}_{j,i})/3\end{aligned}\end{equation}
\begin{equation}\begin{aligned}S_j = \alpha^\top_j h_j,
\end{aligned}\end{equation}
where $\alpha^C_{j,i} $, $\alpha^T_{j,i}$, and $\alpha^{Sl}_{j,i}$ are the attention weight vectors from claim, title and self-attention at the word level. $S_j$ is the formed sentence representation after overall attention for the $j^{th}$ sentence. At the sentence level, we set:
\begin{equation}
\begin{aligned}
\alpha^S_j = (\alpha^C_{j} + \alpha^T_{j} + \alpha^{Sl}_{j})/3\end{aligned}\end{equation}
\begin{equation}
\begin{aligned}doc = \alpha^\top_j h^S,
\end{aligned}
\end{equation}
where $\alpha^C_{j} $, $\alpha^T_{j}$, and $\alpha^{Sl}_{j}$ are the attention weight vectors from claim, title, and self-attention at the sentence level, and $doc$ is the formed document representation after overall attention.

\textbf{Attention on top of Attention (Atop).} Although the fusion of the three kinds of attention weights as an average of them works well, we realize that we lose some context by averaging. To deal with this issue, we use a novel attention on top of attention (Atop) method. We concatenate all three kinds of attentions $\alpha_{con}$ and $\alpha^S_{con}$ at both the word and sentence levels correspondingly. We apply a $\tanh$ activation based dense layer as a scoring function and subsequently, a softmax layer to compute attention weights for each of three kinds of attention:
\begin{equation}
  \begin{aligned}
  \mbox{At word level:}\quad 
  \alpha_{con} = (\alpha^C_{j,i} \mathbin\Vert \alpha^T_{j,i} \mathbin\Vert \alpha^{Sl}_{j,i})\\
  u_{wa} = \tanh(W_{wa} \alpha_{con}  + b_{wa}) \\
  \beta^w= \textrm{softmax}(u_{wa})\\
  S_j = \beta^w_1\alpha^C_{j,i} + \beta^w_2 \alpha^T_{j,i} +  \beta^w_3 \alpha^{Sl}_{j,i}\\
  \mbox{At sentence level:}\quad 
  \alpha^S_{con} = (\alpha^C_{j} \mathbin\Vert \alpha^T_{j} \mathbin\Vert \alpha^{Sl}_{j})\\
  u_{sa} = \tanh(W_{sa} \alpha^S_{con}  + b_{sa}) \\
  \beta^s= \textrm{softmax}(u_{sa})\\
  doc = \beta^s_1\alpha^C_{j} + \beta^s_2 \alpha^T_{j} +  \beta^s_3 \alpha^{Sl}_{j},
  \end{aligned}
\end{equation} 
where $\beta^w$ and $\beta^s$ are the learned attention weight vectors for three kinds of attentions at the word and sentence levels, and $doc$ is the formed document representation after Atop attention.

\textbf{Prediction and Optimization.} We use the overall document representation $doc$ in a softmax layer for the classification. To train the model, we use standard softmax cross-entropy with logits as a loss function, we compute $\hat{y}$, the predicted label as:\begin{equation}
\begin{aligned}
\hat{y} = \textrm{softmax}(W_{cl} doc + b_{cl}).
\end{aligned}
\end{equation}


\subsection{Generating Explainable Summary}

Recent works retrieve documents from the Web as external evidence to support
or refute the claims and thereafter extract  snippets as explanations to
model's decision~\cite{sadhan9,popat2018declare}. However, the extracted
snippets from these methods are often redundant and lack topical diversity. The objective of our summarization algorithm
is to provide ranked list of sentences that are: novel, non-redundant, and
diverse across the topics identified from the text of the documents. In this 
section, we outline the method we utilize for achieving
this objective. 

\textbf{Multi-topic Sentence Model:} Each sentence in the document that is
retrieved against the claim is modeled as a collection of topics: 
$s = \langle a^{(1)}, a^{(2)}, \ldots a^{(k)} \rangle$. Let  $\mathcal{A}$ be the set of topics $a_i\in\mathcal{A}$ across all
candidate sentences from all the pseudo relevant set of documents $D$ for the claim.  

\textbf{Objective}. We formulate the summarization task as a diversification objective. Given a set of relevant sentences $\mathcal{R}$ which are
attended by Atop attention in \textsc{sumo} while establishing the claim's correctness. We have to
find the \emph{smallest} subset of sentences $\mathcal{S} \subseteq \mathcal{R}$
such that \emph{all} topics  $a_i\in\mathcal{A}$ are covered. This
is a variation of the Set Cover problem~\cite{rakesh,korte-book,vazirani-book,williamson-book,set-cover-1,set-cover-2,set-cover-3}. However, unlike
IA-Select~\cite{rakesh} we do not choose to utilize the Max
Coverage variation of the Set Cover problem. Instead, we formulate
it as Set Cover itself~\cite{korte-book,vazirani-book}. That is,
given a set of topics $\mathcal{A}$, find a minimal set of sentences
$\mathcal{S} \subseteq \mathcal{R}$ that cover those
topics~\cite{vazirani-book}. Additionally, the inclusion of each sentence in the
subset $\mathcal{S}$ has a \emph{cost} associated with it, given by:
\begin{equation}
\begin{aligned}
cost(s) = (Score)^{-1}\\
Score = (\lambda {\theta}_s + (1-\lambda) ( W_{wa} + W_{sa})), 
\end{aligned}
\end{equation}
where ${\theta}_s$ is the topic distribution score for sentence $s$ computed
using a topic model (e.g., Latent Dirichlet Allocation~\cite{blei}),
$W_{wa}=\sum^l_{i=1}W_{wa}(i) $ is the average of attention weights of the
words contained in sentence $s$, $W_{sa}$ is the attention weight of the
sentence $s$, and $\lambda$ is a parameter to be tuned. We briefly describe
our adaptation of the Greedy algorithm, which provides an approximate
solution to the Set Cover problem, based on the discussion
in~\cite{korte-book,vazirani-book,williamson-book,set-cover-1,set-cover-2,set-cover-3}.

\RestyleAlgo{boxruled}

\SetAlgoCaptionSeparator{:}

\renewcommand\AlCapFnt{\small\itshape}
\renewcommand\AlCapNameFnt{\small\itshape}
\begin{algorithm}[htb]

    \small
    \SetAlgoLined
    \KwIn{$\mathcal{A}$: Set of topics learned from the topic model for diversification.}
    \myinput{$\mathcal{R}$: Set of sentences, attended by Atop.}
    \KwOut{$\mathcal{S}\subseteq \mathcal{R}$: Diversified set of sentences over $\mathcal{A}$}
    $\mathcal{S} \:\:\gets \phi$ \tcp*{$\mathcal{S}$ contains diversified sentences}
    $\mathcal{A}' \gets \phi$ \tcp*{$\mathcal{A}'$ contains topics covered by $\mathcal{S}$}

    \While{$\mathcal{A}' \neq \mathcal{A}$}{
        \tcc{identify the sentence that covers the most topics and is highly relevant for fact-checking}
        $s^{*} \gets \argmin\limits_{s\in\mathcal{R}\SLASH\mathcal{S}} \frac{cost(s)}{|\mathcal{A} - \mathcal{T'}|}$\;
        $\mathcal{A}' \gets \mathcal{A}' \cup \{a_{s^*}\}$ \tcp*{$a_{s^*}$ is the dominant topic of sentence $s^{*}$}
        $\mathcal{S}  \gets \mathcal{S} \cup s^*$        
    }\caption{Adaption of the approximate Greedy algorithm for Set Cover problem from~\cite{korte-book,vazirani-book,williamson-book,set-cover-1,set-cover-2,set-cover-3} to our topical diversification problem setting. At each iteration, a sentence is chosen that covers the most number of topics reflected by topic distribution score and has the highest attention weights. As an output, we are assured a non-redundant, novel, and a diversified set of sentences.}
\end{algorithm}


\begin{table}[t!!!]
  \centering
  \caption{Dataset Statistics}
  \centering
  \begin{minipage}{.50\textwidth}
  \centering
         \scalebox{0.80}{
         \begin{tabular}{p{3cm} p{2cm} p{2cm}} 

                 \toprule
                 \multicolumn{3}{c}{\textsc{public datasets}}\\
                \midrule
                 \bf \sc statistics    &\bf \sc politifact & \bf \sc snopes\\
           \midrule
                   \#\textsc{claims} &3568 & 4341\\
                \#\textsc{documents}  &29556&29242\\
                   \#\textsc{domains} &3028&3267\\
                   \bottomrule
        \end{tabular}
        }
        \end{minipage}
\begin{minipage}{.50\textwidth}
        \centering
        \scalebox{0.80}{
            \begin{tabular}{p{3cm} p{2cm} p{2cm}} 
            
                 \multicolumn{3}{c}{\textsc{new datasets}}\\
               \midrule
            \bf \sc statistics    &\bf \sc climate & \bf \sc health\\
               \midrule
                   \#\textsc{claims} &104 & 100\\
                 \#\textsc{documents} &1050&978\\
                   \#\textsc{domains} &97&83\\
                   \bottomrule
        \end{tabular}}
   \end{minipage}
   \label{table:datasets}
  \end{table}
  \begin{figure*}[t]
          \centering
          \frame{\includegraphics[width=2\columnwidth]{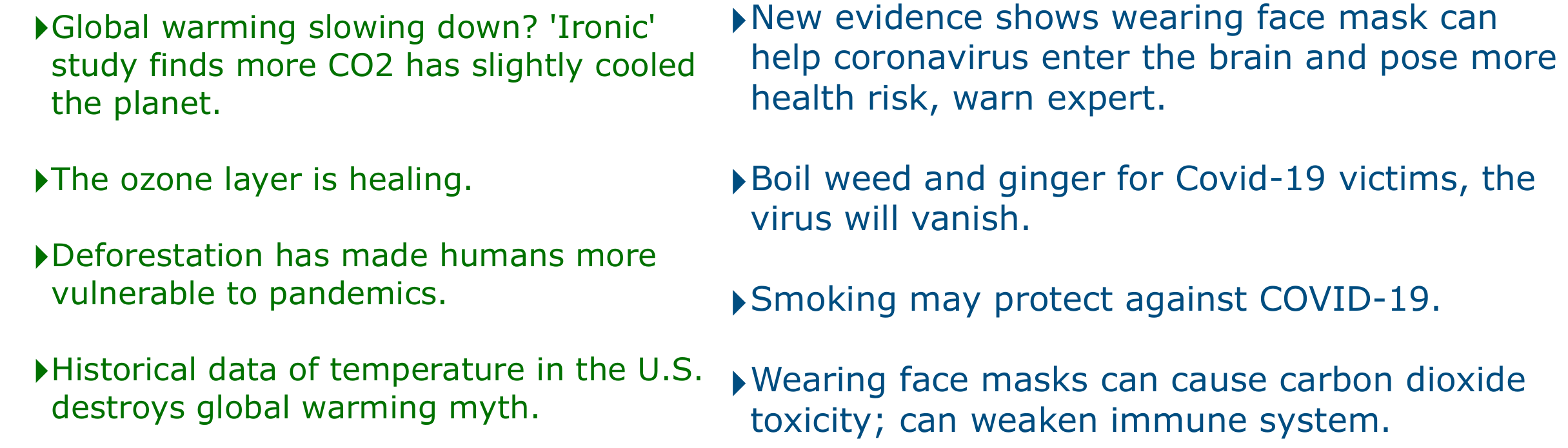}}
          \caption{Examples from climate change and health care dataset}
    \label{fig:examples}
  \end{figure*}

\section{Evaluation}\label{sec:datasets}

\textbf{Datasets.} We use two publicly available datasets, namely
PolitiFact political claims dataset and Snopes political
claims dataset \cite{popat2018declare} for evaluating \textsc{sumo}'s
capability for fact checking. Dataset statistics for both the datasets are
shown in Table~\ref{table:datasets}.  In the case of Politifact,
claims have one of the following labels, namely: ‘true’, ‘mostly true’, ‘half true’,
‘mostly false’, ‘false’, and ‘pants-on-fire,’. We convert ‘true’, ‘mostly true’, and
‘half true’ labels to the ‘true’ and the rest of them to ‘false’ label. For the Snopes dataset, each claim has either ‘true’ or ‘false’ as a
label.

We evaluate \textsc{sumo} for the task of summarization on
PolitiFact, Snopes, Climate, and Health
datasets. The two new datasets, Climate and Health, are
about  climate change and health care respectively. We test \textsc{sumo} only
on the PolitiFact and Snopes dataset for the task of fact
checking as they are magnitudes larger than the new datasets that we release.
The climate change dataset contains claims broadly related to climate change
and global warming from \texttt{climatefeedback.org}. We use each claim as a
query using Google API to search the Web and retrieve external
evidences in the form of search results. Similarly, we create a dataset
related to health care that additionally contains claims pertaining to the
current global COVID-19 pandemic from \texttt{healthfeedback.org}. Examples of
claims from these two datasets are shown in Figure~\ref{fig:examples}. We make the new datasets, publicly available to the research community at the following URL:
\url{https://github.com/rahulOmishra/SUMO/}.

\textbf{\textsc{sumo} Implementation.} We use TensorFlow to implement
\textsc{sumo}. We use per class accuracy and macro F$_1$ scores as performance
metrics for evaluation. We use bi-directional Gated Recurrent Unit (GRU) with
a hidden size of 200, word2vec \cite{Mikolov}, and GloVe
\cite{Pennington14glove:global} embeddings with embedding size of 200 and
softmax cross-entropy with logits as the loss function. We keep the learning rate
as 0.001, batch size as 64, and gradient clipping as 5. All the parameters
are tuned using a grid search. We use $50$ epochs for each model and apply
early stopping if validation loss does not change for more than 5 epochs. We
keep maximum sentence length as 45 and maximum number of sentences in a document as 35.
For the task of summarization, we use Latent Dirichlet Allocation
(LDA) \cite{blei} as a topic model to compute topic distribution scores and
the dominant topic for each candidate sentence.

\section{Results}\label{sec:results}

\subsection{Setup for the Task of Claim Correctness} We experiment with five variants of our proposed 
\textsc{sumo} model and compare with six state-of-the-art methods. The six state-of-the-art methods are as follows. First, we have the basic Long Short Term Memory (LSTM)~\cite{lstm}) 
unit which is used with claim and document contents for classification. Second, we have  
a convolutional neural network (CNN)~\cite{kim2014convolutional} for document classification. 
Third, we compare against the model proposed in~\cite{tang-etal-2015-document} that uses a 
hierarchical representation of the documents using hierarchical LSTM units (Hi-LSTM).
Fourth, we compare against the model proposed in~\cite{yang2016hierarchical} that uses a 
hierarchical neural attention on top of hierarchical LSTMs (HAN) to learn better representations 
of documents for classification. Fifth, we compare against the model proposed in~\cite{popat2018declare} 
that uses a claim guided attention method (DeClarE) for correctness prediction of claims in the 
presence of external evidences. Sixth and finally, we compare against the recent work~\cite{sadhan9} that 
improves on DeClarE method by using latent aspects (speaker, topic, or domain) based attention.

The proposed five variants of our method \textsc{sumo} are as follows. First, we have the 
\textsc{sumo}-AW2V  variant that corresponds to the basic \textsc{sumo} model with 
word2vec embeddings. Second, we have \textsc{sumo}-AtopW2V variant consists of the 
\textsc{sumo} model with \textsc{word2vec} embeddings. Furthermore, in \textsc{sumo}-AtopW2V 
we use Atop method of attention fusion rather than a simple average. Third, we have the 
\textsc{sumo}-AGlove variant, which is the basic \textsc{sumo} model that uses GloVe 
embeddings. Fourth, we have the \textsc{sumo}-AtopGlove variant, that consists of the 
\textsc{sumo} model with GloVe embeddings. Moreover, in \textsc{sumo}-AtopGlove, 
we use Atop method of attention fusion rather than a simple average. Fifth and finally, we have 
the \textsc{sumo}-AtopGlove+source-Emb variant that is similar to \textsc{sumo}-AtopGlove however
with additional source embeddings (domains of retrieved documents). 

\begin{table*}[t!!!]
    \caption{Comparison of the proposed models with various state of the art baseline models for two publicly available datasets.}
\begin{minipage}{0.5\textwidth}
 \centering
\scalebox{.63}{
\setlength{\tabcolsep}{1\tabcolsep}
      \begin{tabular}{c c c c}
      
        \toprule
        \multicolumn{4}{c}{\textsc{PolitiFact}}\\
        \midrule
        \bf Model & \bf True Accuracy & \bf False Accuracy & \bf  Macro F$_1$\\
\midrule         
LSTM &53.51&56.32& 57.89\\
           CNN &55.92& 57.33 &59.39\\
           HAN &60.13& 65.78& 63.44\\
           DeClarE (full)&68.18 &66.01 &67.10 \\
          SADHAN-agg&68.37 &78.23& 75.69\\
          \midrule 
           \textsc{sumo}-AW2V &67.30& 69.22& 70.74\\
           \textsc{sumo}-AtopW2V&67.81& 70.09& 71.15\\
           \textsc{sumo}-AGlove&68.03& 72.57& 72.39\\
          \textsc{sumo}-AtopGlove&68.93& 73.43& 72.79\\
        \textsc{sumo}-AtopGlove+source-Emb&\textbf{69.33}& \textbf{80.08}& \textbf{77.69}\\
                                    \bottomrule
    \end{tabular}
    }
    \end{minipage}%
\begin{minipage}{0.45\textwidth} 
        \scalebox{.63}{
       \setlength{\tabcolsep}{1\tabcolsep}
      \begin{tabular}{cccc}
      \toprule
        \multicolumn{4}{c}{\textsc{Snopes}}\\
      \midrule
        \bf Model & \bf True Accuracy & \bf False Accuracy & \bf  Macro F$_1$\\
     \midrule
           LSTM &69.23& 70.67& 69.89\\
           CNN &72.05& 74.29& 72.63 \\
           HAN &72.89& 76.25& 73.84\\
           DeClarE (full)&60.16 &80.78 &70.47 \\
           SADHAN-agg&79.47 &84.26 &80.09 \\
           \midrule 
           \textsc{sumo}-AW2V &77.32& 80.67& 75.56\\
           \textsc{sumo}-AtopW2V&78.02& 81.66& 76.86\\
           \textsc{sumo}-AGlove&78.74& 82.03& 77.22\\
          \textsc{sumo}-AtopGlove&78.89& 82.46& 78.45\\
                                    \textsc{sumo}-AtopGlove+source-Emb&\textbf{81.29}& \textbf{86.82}& \textbf{82.93}\\
\bottomrule
    \end{tabular}
    }
    \end{minipage}
    \label{tab:results}
  \end{table*}%
\subsection{Claim Correctness Task Results} The results for establishing claim correctness are shown in Table 2. We observe that the basic LSTM based model achieves $57.89\%$ and $69.89\%$ in terms of macro F$_1$ accuracy 
in prediction of claim correctness for \textsc{Politifact} and \textsc{Snopes}, respectively. The CNN model performs 
slightly better than LSTM as it captures the local contextual features better. The hierarchical 
attention network outperforms CNN with macro F$_1$ accuracy of $63.4\%$ and $73.84 \%$. The reason for 
this improvement is hierarchical representation using word and sentence level attention. The state of 
the art DeClarE model provides significant improvements on baseline methods with macro F$_1$ accuracy of 
$67.10\%$ and $70.47 \%$. This gain can be attributed to claim guided attention and source embeddings. 
However, we observe that this design of claim based attention is not very effective. The more recent work,
SADHAN improves on DeClarE, which uses a similar design for claim-oriented attention 
and incorporates a more comprehensive structure by using several latent aspects to guide attention. 

SADHAN outperforms DeClarE with macro F$_1$ accuracy of $75.69\%$ and $80.09 \%$, respectively. Interestingly, 
we observe that the basic \textsc{sumo} model with word2vec embeddings performs better than 
DeClarE with source embeddings. This observation is a clear indication of the superiority of our claim- 
and title-driven attention design. The \textsc{sumo} with Atop attention fusion is more effective than 
a simple average fusion of attention weights, which becomes apparent from the gain in macro F$_1$ accuracy 
in both the datasets. \textsc{sumo} with pertained GloVe embeddings outperforms the 
word2vec versions of \textsc{sumo} as the GloVe embeddings are trained on a large corpus and therefore captures better context for the words. \textsc{sumo}-AtopGlove+source-Emb outperforms 
all the other models and it is statistically significant with a p-value of $2.79 \times 10^{-3}$ for \textsc{PolitiFact} 
and $3.09 \times 10^{-4}$ for \textsc{Snopes}. The 
statistical significance values were computed using a two sample Student's t-test. 
We notice that \textsc{sumo} could not outperform SADHAN without source embeddings, as SADHAN uses 
the very complex structure, having three parallel models with hierarchical latent aspects guide attention. 
However, SADHAN has many drawbacks. First, it is challenging to train and requires more hardware resources 
and time. Second, the latent aspects are not available for all the Web claims. Therefore, it is not generalizable. 
Third, it fails to accommodate new values of latent variables at the test time. 

\subsection{Setup for the Task of Summarization}
For the evaluation of the summarization capability of \textsc{sumo}, we create gold reference
summaries for claims. For
creating the gold reference summaries, we include all the facts related to the 
claim, which are important for the claim correctness prediction,
non-redundant, and topically diverse. We find that the descriptions provided for a
claim on fact-checking websites such as \texttt{snopes.com} and
\texttt{politifact.com} are suitable for this purpose.
We use cosine similarity score of 0.4 between claims and sentences of
description to filter out irrelevant or noisy sentences. As evaluation metrics, we use
ROUGE-1, ROUGE-2, and ROUGE-L scores. The ROUGE-1 score represents the overlap of
unigrams, while  the ROUGE-2 score represents the overlap of bigrams between the
summaries generated by the \textsc{sumo} system and gold reference summaries.
The ROUGE-L score measures the longest matching sequence of words using Longest Common
Sub-sequence algorithm.

Standard summarization techniques are not useful in such a scenario as the
objective of summarization with standard techniques is usually not
fact-checking. Hence, we compare the \textsc{sumo} results with an information retrieval (BM25) and a natural language processing based method (QuerySum). BM25 is a ranking function,
which uses a probabilistic retrieval framework and ranks the documents based
on their relevance to a given search query. We use Web claims as a query and
apply BM25 to get the most relevant sentences from all the documents retrieved
for the claim. We also compare the results with the query-driven attention based
abstractive summarization method QuerySum~\cite{nema-etal-2017-diversity},
which also uses a diversity objective to create a diverse summary. We use
ROUGE metrics with a gold reference summary to evaluate the generated
summaries. 
\subsection{Comparison of Summarization Results} Results for the task of summarization are shown in Table 3, the QuerySum method performs significantly better than BM25 with
a ROUGE-L score of 30.16 as it uses query-driven attention and diversity
objective, which results in a diverse and query oriented summary. The proposed
model \textsc{sumo} outperforms QuerySum with a ROUGE-L score of 35.92. We
attribute this gain to the use of word and sentence level weights, which are
trained using back-propagation with correctness label. We also notice that in
QuerySum some sentences are related to the claim but are not useful for
fact checking. Therefore, they are absent in the gold reference summary. The results for \textsc{sumo} are statistically significant ($p$-value $ = 1.39\times10^{-4}$) using a pairwise Student's t-test.  
   
\begin{table}[t!!!]
    \centering
    \caption{Results for the Task of Summarization.}
    \centering
    \scalebox{.75}{
    \setlength{\tabcolsep}{1.25\tabcolsep}
    \begin{tabular}{r r r r}
\toprule
        \bf Model &\bf ROUGE-1 & \bf ROUGE-2&\bf ROUGE-L \\
        \midrule
           BM25 &26.08 &14.78&29.98\\
        QuerySum &29.78&16.49&30.16\\
           \textsc{sumo}&\textbf{33.89}&\textbf{19.21}&\textbf{35.92}\\
           \bottomrule
    \end{tabular}
    }
  \end{table}

\section{Conclusion}\label{sec:conclude}
We presented \textsc{sumo}, a neural network based approach to generate explainable and topically diverse summaries for verifying Web claims. \textsc{sumo} uses an improved version of hierarchical claim-driven attention along with title-driven and self-attention to learn an effective representation of the external evidences retrieved from the Web. Learning this effective representation in turn assists us in establishing the correctness of textual claims. Using the overall attention weights from the novel Atop attention method and topical distributions of the sentences, we generate extractive summaries for the claims. In addition to this, we release two important datasets pertaining to climate change and healthcare claims. 

In future, we plan to investigate the BERT \cite{devlin-etal-2019-bert} and other Transformer \cite{Vaswani} architecture based embedding methods in place of GloVe \cite{Pennington14glove:global} embeddings for better contextual representation of words.  
\bibliographystyle{acl_natbib}

\balance
\end{document}